\documentclass{bmvc2k}

\title{Depth-aware Object Segmentation and Grasp Detection for Robotic Picking Tasks}

\addauthor{Stefan Ainetter}{stefan.ainetter@icg.tugraz.at}{1}
\addauthor{Christoph Böhm}{christoph.boehm@ieee.org}{2}
\addauthor{Rohit Dhakate}{rohit.dhakate@ieee.org}{2}
\addauthor{Stephan Weiss}{stephan.weiss@ieee.org}{2}
\addauthor{Friedrich Fraundorfer}{fraundorfer@icg.tugraz.at}{1}
\addinstitution{
 Institute of Computer Graphics \& Vision\\
 Graz University of Technology\\
 Graz, Austria
 }
\addinstitution{
 Institute of Smart System Technologies\\
 University of Klagenfurt\\
 Klagenfurt, Austria 
}

\runninghead{Ainetter et al.}{Depth-aware Obj. Seg. \& Grasp Det. for Robotic Picking}

\usepackage{pifont}
\usepackage{multirow}
\newcommand{\cmark}{\ding{51}}%
\newcommand{\xmark}{\ding{55}}%

\begin{document}
\maketitle
\vspace{-1em}
\begin{abstract}
In this paper, we present a novel deep neural network architecture for joint class-agnostic object segmentation and grasp detection for robotic picking tasks using a parallel-plate gripper. We introduce depth-aware Coordinate Convolution (CoordConv), a method to increase accuracy for point proposal based object instance segmentation in complex scenes without adding any additional network parameters or computation complexity. Depth-aware CoordConv uses depth data to extract prior information about the location of an object to achieve highly accurate object instance segmentation. These resulting segmentation masks, combined with predicted grasp candidates, lead to a complete scene description for grasping using a parallel-plate gripper. We evaluate the accuracy of grasp detection and instance segmentation on challenging robotic picking datasets, namely Siléane and OCID\_grasp, and show the benefit of joint grasp detection and segmentation on a real-world robotic picking task.
\end{abstract}
\vspace{-2em}
\section{Introduction}
\label{sec:intro}
Automated grasping in challenging environments is one of the most active fields of research in robotics. In order to successfully grasp and move objects, accurate information about potential grasp candidates, which are areas where the gripper can grasp the object, as well as shape and location of these objects are highly important. Previous works used deep neural networks to predict highly accurate grasp candidates for parallel-plate grippers with one or multiple objects in the scene. Recent works focused on additional object detection \cite{yang2019task,zhang2018multi,park2020multi} or semantic segmentation \cite{ainetter2021}. This additional information is used to identify and pick specific objects or a class of objects, and to understand the relation between objects in a cluttered scene. 
\\Other researchers focused on class-agnostic object instance segmentation \cite{xie2020uois3d,ito2020point}, arguing that pixel-accurate instance segmentation of unknown objects is highly important for robotic picking, as it helps to understand each object's shape and location accurately and can be used to avoid potential collisions during grasping (as shown in \cite{murali20206}). Although these methods provide highly accurate results for instance segmentation, they do not directly provide grasp candidates. \\ In this work, we address joint grasp detection for parallel-plate grippers as well as class-agnostic instance segmentation, by proposing an end-to-end trainable multi-task deep neural network. Our network provides high quality results for grasp detection and pixel-wise instance segmentation for multiple graspable objects in complex scenes, where objects highly overlap. 
We achieve this, by introducing our novel depth-aware CoordConv module. Depth-aware CoordConv enables us to generate additional feature maps which encode a location prior related to the searched object instance, which are used to predict accurate object segmentation masks. We evaluate our method on publicly available datasets for robotic manipulations with high complexity. Additionally, we demonstrate our method on a real-world robotic picking task, and show simple ways to use the predicted instance segmentation to enhance grasp accuracy.  
The main contributions of our paper are the following:
\begin{enumerate}
    \item An end-to-end trainable deep neural network architecture for joint grasp detection and class-agnostic instance segmentation, which yields state-of-the-art performance for grasp detection and provides valuable information for scene understanding in robotic picking tasks.
    \vspace{-6pt}
    \item Depth-aware CoordConv, a novel method to increase class-agnostic instance segmentation accuracy in highly challenging scenes with multiple overlapping objects. Notably, it does not introduce additional network parameters or computation complexity.%
    \vspace{-6pt}
    \item Extensive experiments including public grasp datasets and real-world experiments, to show the benefit of using joint grasp detection and instance segmentation in scenes with high clutter and occlusions. By leveraging the predicted segmentation masks we show how to optimize the picking and placing of target objects in a real environment.
\end{enumerate}
\vspace{-18pt}
\section{Related Work}
Grasping is a widely studied field in robotics \cite{bohg2013data,bicchi2000robotic,caldera2018review}. A popular approach to solve it is Reinforcement Learning \cite{quillen2018deep,james2019sim,kalashnikov2018scalable}, which uses feedback of grasp attempts for learning in real or simulated environments. Regarding supervised learning approaches, we can distinguish between planar and spatial grasp detection. Methods for spatial grasp detection \cite{mousavian20196,murali20206,gualtieri2018learning} usually predict 6-DOF grasp poses, whereas planar grasp detection represents grasp candidates as oriented rectangles or points in the image. In the following, we will focus our comparison on methods similar to ours using planar grasp detection.
\\\textbf{Planar Grasp Detection.} First approaches using deep neural networks and supervised learning to predict multiple grasp candidates for a single object are \cite{redmon2015real,lenz2015deep}. Latest research focuses on detection of grasp candidates for multiple objects \cite{pinto2016supersizing,kumra2017robotic,guo2017hybrid,zhou2018fully,asif2019densely,morrison2020learning}. The works of \cite{yang2019task,zhang2018multi} perform grasp detection with additional object detection to identify the relationship between multiple objects in complex scenes.
\\ \textbf{Object segmentation for robotic picking tasks.}
The authors of \cite{xie2020uois3d} provide a method for unseen object instance segmentation for robotic environments by using an encoder/decoder CNN architecture with clustering based on Hough voting. The authors of \cite{danielczuk2019segmenting} predict class-agnostic segmentation of objects from depth images using a variant of Mask R-CNN \cite{he2017mask}. However, region proposal based segmentation methods like Mask R-CNN often fail to perform well in complex scenes where objects highly overlap, which is often the case for robotic picking. \cite{ito2020point} proposed a method for point proposal based instance segmentation for robotic picking tasks. To generate these point proposals, \cite{ito2020point} added specific network branches, which introduce additional overhead in terms of network parameters and computation cost. Our method, which also relies on point proposals for instance segmentation, compensates this disadvantage by directly utilizing predicted grasp candidates as point proposals for instance segmentation. 
\\ \textbf{Joint grasp detection and segmentation.}
The authors of \cite{araki2020mt} proposed a network for semantic segmentation, and grasp detection represented as points for a suction cup manipulator using multi-task learning with a single deep neural network.
Other recent work \cite{ainetter2021} proposed a method for joint grasp detection and class-specific semantic segmentation. However, semantic segmentation is not sufficient to distinguish between multiple instances of the same object class.
\vspace{-12pt}
\section{Problem Statement}
We propose an end-to-end trainable pipeline to detect reliable grasp candidates for robotic grasping, by combining grasp detection with semantic and class-agnostic instance segmentation. A brief introduction to the tasks of grasp detection and segmentation is given below:
\\\textbf{Grasp detection.}
We use the five-dimensional bounding box representation \cite{lenz2015deep} to define grasp candidates for robotic picking. A grasp candidate $\mathbf{g}$ is defined as
\begin{equation}
\mathbf{g}= (x,y,w,h,\theta),
\label{eq:grasp_rep_init}
\end{equation}
where $(x,y)$ define the center of the grasp candidate, $w$ and $h$ define the width and height of the bounding box, and $\theta$ defines the orientation of the grasp candidate.
\vspace{2pt}
\\ \textbf{Semantic segmentation.} Semantic segmentation is the task of classifying each pixel of an input image into one of the predefined class labels. For grasp detection of unknown objects, we define the set of semantic class labels as $l \in \{1,...,N\}$, whereas $N=2$ is the number of classes for foreground/background segmentation.
\\\textbf{Class-agnostic instance segmentation.}
Class-agnostic (or category-agnostic) instance segmentation is the task of predicting pixel-wise segmentation masks for each object instance in the scene. Class-agnostic means that no information about object classes is available, for example the objects are either not seen or not known during training.
\begin{figure}[htp!] 
      \centering
 		\includegraphics[trim={0cm 0cm 0cm 0cm},width=\linewidth]{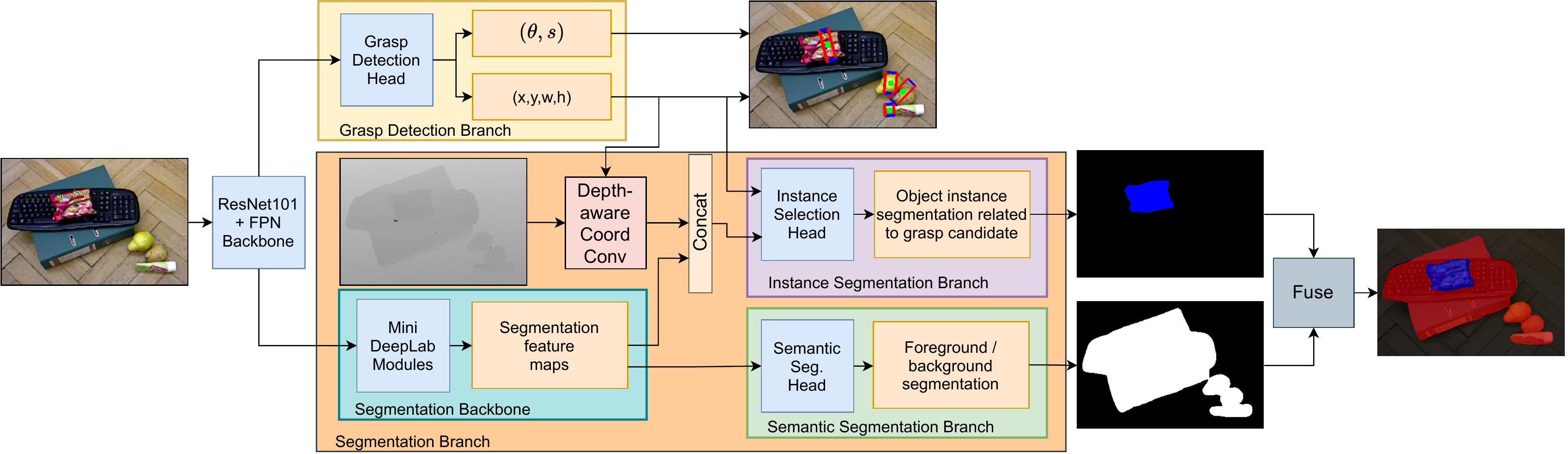}
	\caption[Network architecture]{Architecture of our proposed method. 
	Both branches for grasp detection and segmentation share the backbone network as feature extractor. The output of our network consists of grasp candidates for multiple graspable objects in the scene (green dot denotes center of grasp candidate, red lines denote opening width of gripper, blue lines denote parallel plates). Additionally, the network predicts accurate object instance seg. masks for one or multiple graspable objects and fore-/background segmentation. The final segmentation result shows the object instance mask related to the grasp candidate $\mathbf{g}$ with the highest confidence score $s$ which is highlighted in blue, and all other objects in the scene (which can be seen as obstacles) are red.}
	\label{fig:Det_Seg_Net}
	\vspace{-12pt}
\end{figure}
\vspace{-12pt}
\section{Architecture Overview}
Our proposed architecture for joint grasp detection and segmentation consists of a shared backbone for feature extraction combined with task-specific branches for grasp detection, semantic, and instance segmentation. A ResNet-101 \cite{he2016identity} with Feature Pyramid Network (FPN) \cite{lin2017feature} is used to extract feature maps from an RGB input image, which are then used as input for joint grasp detection and segmentation. Figure \ref{fig:Det_Seg_Net} shows an overview of our network architecture. Hereafter, we describe each component of our network in detail.
\vspace{-6pt}
\subsection{Grasp Detection Branch}
\vspace{-6pt}
The grasp detection branch performs the task of predicting grasp candidates for multiple graspable objects in the scene. To achieve this, we use a region proposal based grasp detection network, similar to \cite{ainetter2021}. 
The grasp detection head takes previously extracted feature maps as input and predicts grasp candidates in form of axis-aligned rectangular boxes and the rotation parameter $\theta$. Additionally, it predicts a confidence score $s \in [0,1]$, whereas higher values indicate the network is more confident that the grasp candidate is valid. The final result is a number of grasp candidates $\mathbf{g}$ (see Eq. \ref{eq:grasp_rep_init}) and corresponding confidence scores $\mathbf{s}$. 
\vspace{-6pt}
\subsection{Segmentation Branch}
The segmentation branch consists of a segmentation backbone, and sub-branches for semantic and instance segmentation. We will now describe all parts in detail.
\vspace{-10pt}
\subsubsection{Segmentation Backbone}
\vspace{-6pt}
The segmentation backbone takes as input a subset of the extracted feature maps from the ResNet101 + FPN backbone network, which correspond to the first four scales of the FPN. Mini-DeepLab modules \cite{porzi2019seamless} are then applied independently to each input. The output of each module consists of 128 feature maps which are down-sampled by the factor of x4 compared to the input image size. The feature maps of all modules are concatenated and the resulting final 512 segmentation feature maps are used as input for both the semantic and instance segmentation branch.
\vspace{-10pt}
\subsubsection{Semantic Segmentation Branch}
\vspace{-6pt}
The semantic segmentation branch assigns a class label $l \in \{1,...,N\}$ to each pixel of the input image, whereas $N=2$ is the number of classes for foreground/background segmentation. Note that if class information is available, the set of semantic classes can be adjusted accordingly. The semantic segmentation head takes as input the feature maps from the segmentation backbone, which are fed to a 1x1 convolution layer with $N$ output channels. The resulting logits are used as input for a final softmax layer which provides class probabilities for each pixel of the input image.
At the end, the probability maps are bilinearly upsampled to the size of the input image which results in the final foreground/background segmentation.  

\vspace{-10pt}
\subsubsection{Instance Segmentation Branch}
\begin{figure}[htp!] 
      \centering
 		\includegraphics[trim={0cm 0.5cm 0cm .5cm},width=0.9\linewidth]{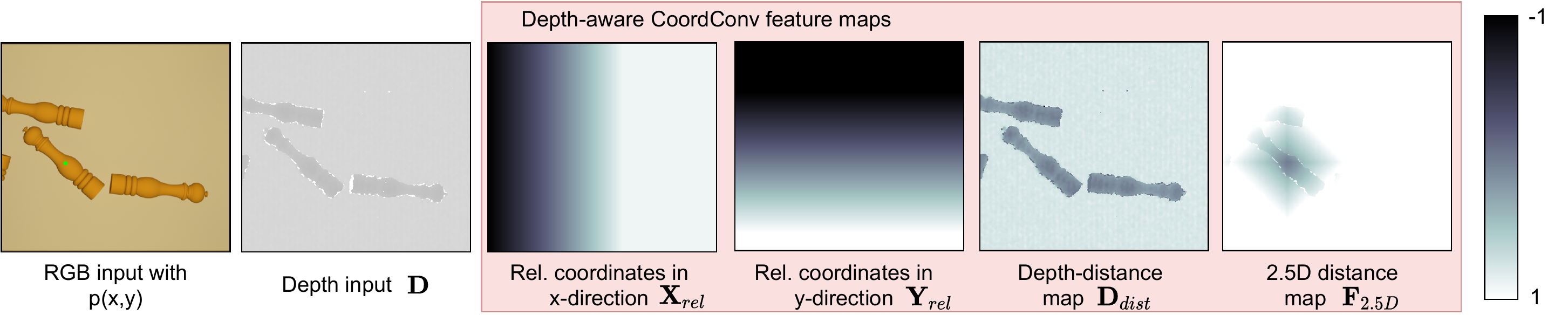}
	\caption[Example DACC]{Visualizations of feature maps calculated using depth-aware CoordConv. The two left images show an RGB input with the point proposal $p(x,y)$ marked as green dot, and the depth input $D$, respectively. Additionally, the corresponding feature maps generated using depth-aware CoordConv are shown. One can see that these feature maps encode a location prior for the selected object.}
	\label{fig:example_dacc}
\end{figure}

\vspace{-6pt}
\label{sec:depth_aware_IS}
Our instance segmentation branch is based on two main components, depth-aware CoordConv and the instance selection head. Both components take a point proposal $p(x,y)$ as input, which is used to select a specific object for instance segmentation. Our full pipeline (shown in Figure \ref{fig:Det_Seg_Net}) utilizes the center of a predicted grasp candidate $\mathbf{g}$ as point proposal. Note that utilizing information about our predicted grasp candidates provides a clear advantage to other point proposal based instance segmentation methods, e.g. \cite{sofiiuk2019adaptis,ito2020point}, which need an additional module (e.g. an additional neural network branch) to predict point proposals.
\\Depth-aware CoordConv calculates feature maps to encode a location prior for a specific object based on $p(x,y)$. The instance selection head performs the task of predicting an instance segmentation mask related to the point proposal $p(x,y)$. A detailed description about both parts follows below:
\\\textbf{Depth-aware CoordConv.} It is based on the idea of coordinate convolution \cite{liu2018coordconv}, which is a mechanism to make a convolutional neural network aware of input coordinates. This counteracts the fact that these networks have the property of translation invariance, producing similar outputs for similar objects at different locations.  \\Depth-aware CoordConv generates several feature maps that encode a positional prior for a specific object instance.
This helps to distinguish multiple object instances in highly stacked scenes, which is often the case for robotic picking tasks. First, we calculate $\mathbf{X}_{rel}$ and $\mathbf{Y}_{rel}$, which encode the relative coordinates (rel. CoordConv) for the point proposal $p(x,y)$ in x and y direction, respectively. Both $\mathbf{X}_{rel}$ and $\mathbf{Y}_{rel}$ are then divided by a hyperparameter $R$, which roughly sets the maximum object size. Similar methods have already been used in the literature, e.g. in \cite{sofiiuk2019adaptis}. 
Additionally, we define the depth-distance map $\mathbf{D}_{dist} = \alpha \cdot (\mathbf{D} - \mathbf{D}(p))$ with $\mathbf{D}_{dist} \in [-\alpha,\alpha]$,
whereas $\mathbf{D} \in [0,1]$ is the normalized depth input and $\alpha$ is a hyperparameter used as scaling factor. At last, we calculate a 2.5D distance map using $\mathbf{F}_{2.5D} = \sqrt{\mathbf{X}_{rel}^2 + \mathbf{Y}_{rel}^2 + \mathbf{D}_{dist}^2}$. All calculated feature maps from depth-aware CoordConv are then clamped to $(-1,1)$. 
Figure \ref{fig:example_dacc} shows an example of these final feature maps.
The output of depth-aware CoordConv is concatenated with the feature maps produced by the segmentation backbone,  and together they are passed to the instance selection head. Note that this usage of depth information does not add any network parameters or computational complexity (compared to other methods that use multiple input modalities to improve performance, e.g. \cite{wang2020makes,loghmani2019recurrent}), and is therefore a simple and effective way to utilize depth information.
\vspace{-6pt}
\vspace{-6pt}
\\\textbf{Instance selection head.} The instance selection head is based on \cite{sofiiuk2019adaptis} and takes as input the segmentation and depth-aware feature maps, as well as the point proposal $p(x,y)$. First, the number of seg. feature maps is reduced by the factor of 2 using a 1x1 convolutional layer. Then, this feature maps are passed to three convolutional layers, followed by an AdaIN layer \cite{huang2017arbitrary} and again two convolutional layers whereas the last one performs 1x1 convolution. Note that the AdaIN layer is used to guide the instance selection head to segment the specific object selected using the point proposal $p(x,y)$. This is done by extracting a descriptive feature vector from the segmentation features at the location $p(x,y)$ using a few fully connected layers, which is then passed to the AdaIN layer.
Each convolutional layer in the instance selection head is followed by ReLU activation and Batch Normalization \cite{ioffe2015batch}, except the 1x1 convolutional layers. The resulting logits are bilinearly upsampled to the size of the input image and then used as input for a softmax function which provides probabilities that a pixel belongs to the selected object instance.
\vspace{-6pt}
\subsection{Loss Function}
\vspace{-6pt}
The loss function for joint grasp detection, semantic and instance segmentation, is defined as $\mathcal{L}= \lambda_{grasp}\mathcal{L}_{grasp} + \lambda_{sem}\mathcal{L}_{sem} + \lambda_{inst}\mathcal{L}_{inst}$,
with the grasp detection loss $\mathcal{L}_{grasp}$, the semantic segmentation loss $\mathcal{L}_{sem}$, and the instance selection loss $\mathcal{L}_{inst}$. The grasp detection loss $\mathcal{L}_{grasp}$, based on \cite{ainetter2021}, is a combination of a regression loss for predicting axis-aligned bounding box parameters $(x,y,w,h)$, and a classification loss for predicting the orientation parameter $\theta$. For $\mathcal{L}_{sem}$ we used a weighted per-pixel log loss similar to \cite{porzi2019seamless}. As instance selection loss $\mathcal{L}_{inst}$ we used a Normalized Focal Loss as described in \cite{sofiiuk2019adaptis}. The specific hyperparameters $\lambda$ are used for loss balancing. Note that the grasp detection branch and segmentation branch are supported with distinct losses, and depth-aware CoordConv influences the grasp detection only indirectly through the shared backbone. We refer to the supplementary material for detailed information about the loss function. 
\vspace{-12pt}
\section{Experiments and Results}
\vspace{-6pt}
To assess the benefits of our proposed method, we divided the evaluation into three parts: 1) evaluation of our instance segmentation method focusing on the benefit of depth-aware CoordConv, 2) evaluation of joint grasp detection and object segmentation on the challenging OCID\_grasp dataset \cite{ainetter2021}, and 3) real-world robotic grasp experiments with the focus on using instance segmentation masks to optimize picking and placing of objects. Note that for all experiments, we initialize the backbone network with pre-trained ImageNet \cite{russakovsky2015imagenet} weights and freeze the parameters of the first two backbone network modules. 
\\\textbf{Data Augmentation.} Throughout all training runs, we use the following method for data augmentation. For each RGB-D input image, we apply random rotation between $0^{\circ} \text{ and } 360^{\circ}$, and randomly translate the input data in x and y direction independently up to $50px$.
\\\textbf{Evaluation metrics.}
The grasp accuracy is measured using the Jaccard index, with the following criteria needed to be true for a valid grasp candidate:
\vspace{-6pt}
\begin{enumerate}
    \item The angle difference between predicted grasp candidate $\mathbf{g}_{p}$ and ground truth grasp candidate $\mathbf{g}_{gt}$ is within $30^{\circ}$ and
    \vspace{-6pt}
    \item the Intersection over Union (IoU) of them is greater than $0.25$. 
\end{enumerate}
\vspace{-6pt}
For instance and semantic segmentation, we calculate the mean IoU between predicted and ground truth segmentation.
\vspace{-6pt}
\subsection{Depth-aware Instance Segmentation in Complex Scenes}
\begin{figure}[htp!] 
      \centering
 		\includegraphics[trim={0cm 0 0cm 0cm},width=0.9\linewidth]{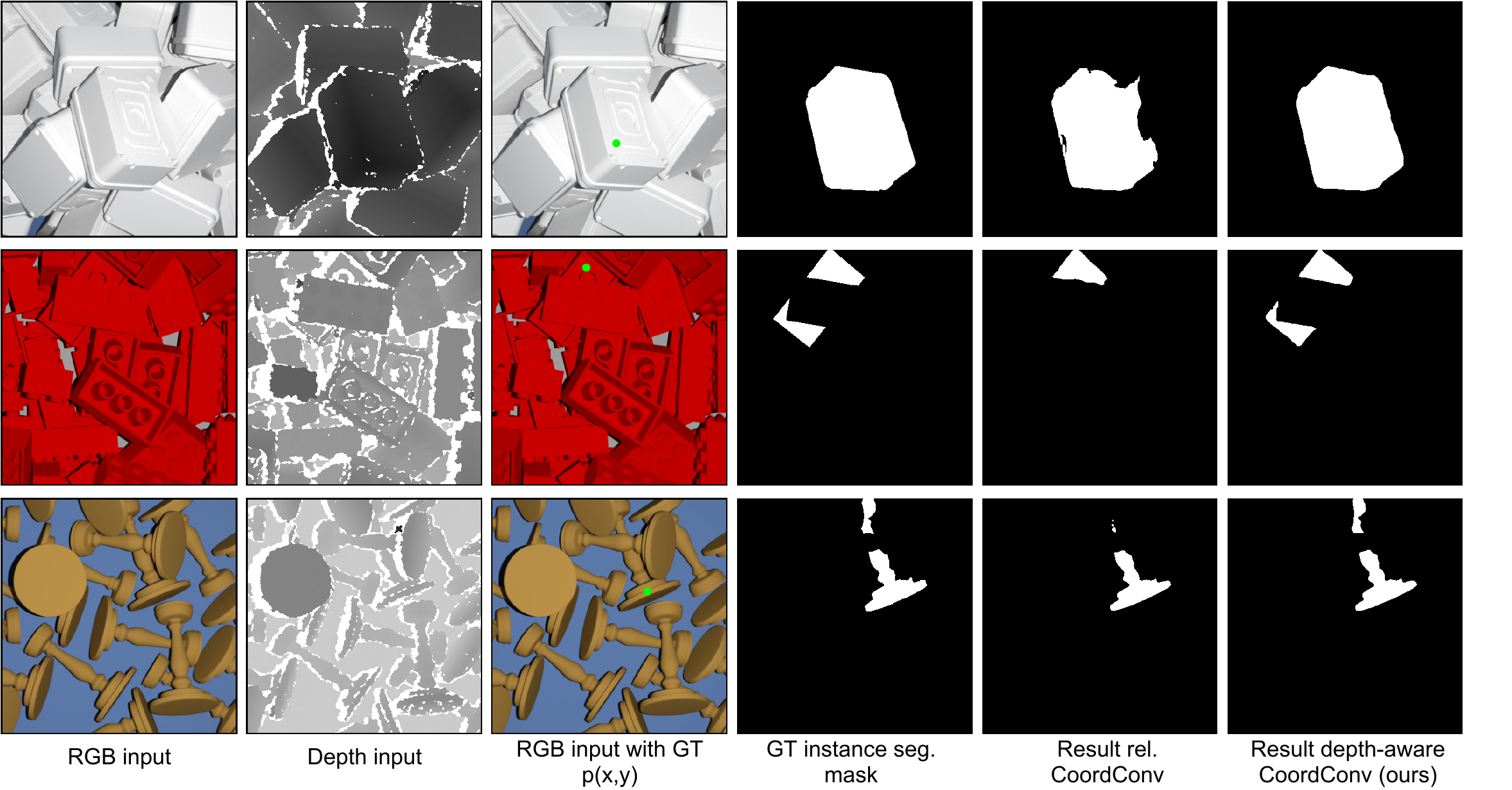}
 	\vspace{-6pt}
	\caption[]{Qualitative results for instance segmentation on Siléane dataset. Note that the point proposals $p(x,y)$ in this experiment are given as ground truth. The resulting instance segmentation masks are only shown for one point proposal to ensure clarity. By comparing the results of rel. CoordConv \cite{sofiiuk2019adaptis} and depth-aware CoordConv (ours), one can see that our method increases segmentation accuracy especially for highly occluded objects.}
	\label{fig:Sileane_results}
	\vspace{-18pt}
\end{figure}

\vspace{-6pt}
To evaluate the benefit of depth-aware CoordConv for object instance segmentation, we used the synthetic part of the Siléane dataset \cite{bregier2017iccv} (excluding images with no objects), which consists of 1594 RGB-D images depicting various numbers of object instances in bulk, with pixel-wise annotated segmentation masks. Figure \ref{fig:Sileane_results} shows examples for RGB and depth data.
\\\textbf{Dataset preparation.}  First, we center crop each image to a size of $400 \times 400px$. 
During training, we randomly sampled point proposals $p(x,y)$ on random object instances, and predicted the corresponding instance segmentation. During testing, we used ground truth point proposals which were sampled in advance, to ensure fair comparison throughout our experiments. 
We performed an 80:20 image-wise split into training and test set, using 80\% of data for training and 20\% for testing. The image-wise split means that images are separated randomly without considering which objects are in the scene.
\\\textbf{Training procedure.} The network for this experiment consists of a ResNet-101 + FPN backbone and the segmentation branch (without the grasp detection branch, see Figure \ref{fig:Det_Seg_Net}). We used SGD as optimizer, with a learning rate of $0.02$ and a weight decay of $0.0001$. We used a batch size of $4$ and randomly sampled $9$ point proposals for each image. For loss balancing, we chose weighting factors $\lambda_{sem}= 1.0$ and $\lambda_{inst}=1.0$.
\\\textbf{Quantitative evaluation.}
Table \ref{tab:sileane_eval} shows the influence of feature maps generated by depth-aware CoordConv. As previously shown in \cite{sofiiuk2019adaptis}, we can confirm that using rel. CoordConv feature maps as location prior increase segmentation accuracy. Furthermore, the results show that our depth-aware CoordConv significantly improves instance segmentation accuracy for complex scenes. We used the same network architecture, hyperparameters and training schedule for all experiments. Therefore, the improvement for instance segmentation is directly related to depth-aware CoordConv. Additional experiments with different variants of depth-aware CoordConv are provided in the supplementary material. 
\begin{table}[htp!]
\centering
\begin{tabular}{|c|c|c|c|c|}
\hline
\begin{tabular}[c]{@{}c@{}} Network\\ Architecture\end{tabular}  & \begin{tabular}[c]{@{}c@{}}Rel.\\ CoordConv \cite{sofiiuk2019adaptis} \end{tabular} & \begin{tabular}[c]{@{}c@{}} Depth-dist.\\ Map (ours)\end{tabular} & \begin{tabular}[c]{@{}c@{}}2.5D Dist.\\ Map (ours)\end{tabular} & \begin{tabular}[c]{@{}c@{}}Inst. Seg.\\IOU (\%)\end{tabular} \\ \hline
Backbone + Seg. Branch & \xmark & \xmark & \xmark & 83.01 \\ \hline
Backbone + Seg. Branch & \cmark & \xmark & \xmark  & 85.63 \\ \hline
Backbone + Seg. Branch & \cmark & \cmark & \cmark   & \textbf{91.27}\\ \hline
\end{tabular}
\vspace{6pt}
\caption{Comparison of different configurations of CoordConv for instance segmentation. The results show that using all feature maps generated by our depth-aware CoordConv module significantly improves segmentation accuracy in complex scenes.}
\label{tab:sileane_eval}
\vspace{-6pt}
\end{table}
\\\textbf{Qualitative evaluation.}
Figure \ref{fig:Sileane_results} shows qualitative results for object instance segmentation with and without depth-aware CoordConv. The results show that using our method significantly improves instance segmentation in scenes where multiple objects highly overlap.

\subsection{Joint Grasp Detection and Object Segmentation}
\label{sec:sileane_exp}
\begin{figure}[htb!] 
      \centering
 		\includegraphics[width=\linewidth]{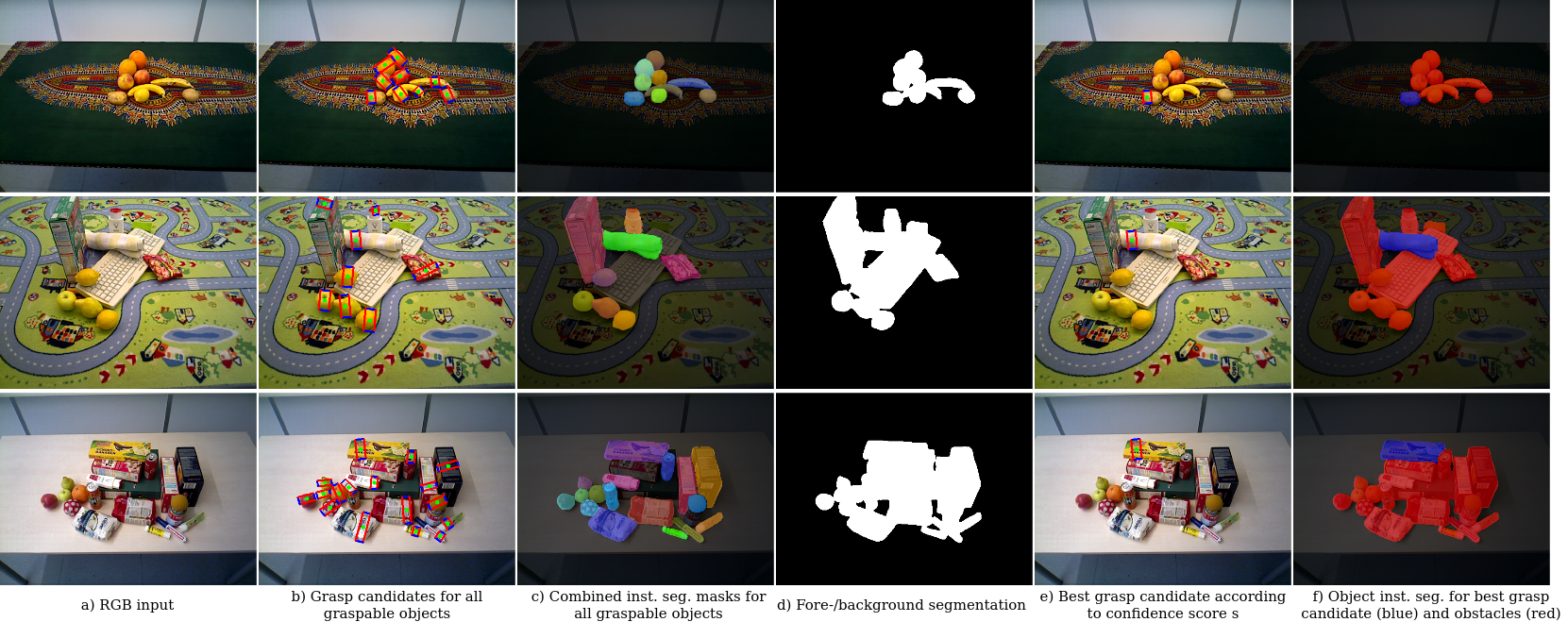}
 		\vspace{-2em}
 		\caption[OCID\_grasp results]{Qualitative results for joint grasp detection and segmentation on OCID\_grasp (using \textit{(ours) with depth-aware CC}). Results in b) - d) are used for quantitative evaluation in Table \ref{tab:OCID_results}. Note that our method is able to accurately detect grasp candidates and corresponding inst. seg. masks for multiple graspable objects in the scene. Each color in c) represents a independent object instance. As one can see in e) and f), for real-world robotic grasping of unknown objects, we are primarily interested in one valid grasp candidate, and a segmentation which defines the graspable object and all other objects in the scene (which are in this case obstacles).}
	\label{fig:OCID_full_results}
	\vspace{-1em}
\end{figure}

\vspace{-6pt}
The recently proposed OCID\_grasp \cite{ainetter2021}, a dataset extension for OCID \cite{suchi2019easylabel}, contains RGB-D images of diverse settings of objects and backgrounds, with hand annotated grasp candidates and object instance segmentation masks as ground truth. Figure \ref{fig:OCID_full_results} a) shows sample RGB images of the dataset. To highlight the effect of depth-aware CoordConv, we evaluated the full network architecture for joint grasp detection and segmentation (as proposed in Figure \ref{fig:Det_Seg_Net}) with and without  depth-aware CoordConv.
\\ \textbf{Training schedule.}
The network was trained end-to-end using a learning rate of 0.02 with weight decay of 0.0001, a momentum factor of 0.9 with enabled nesterov momentum and SGD as optimizer. We used a batch size of 3 during training and weighting factors of $\lambda_{grasp}= 1.0$, $\lambda_{sem}=0.7$ and $\lambda_{inst}=0.7$ for loss balancing. We performed an 80:20 image-wise split into training and test set, using 80\% of data for training and 20\% for testing. During training, we randomly sampled point proposals for object instance segmentation, during evaluation, we used the center of the grasp candidates $\mathbf{g}$ as point proposals $p(x,y)$.
\\ \textbf{Quantitative \& qualitative results.} Table \ref{tab:OCID_results} shows quantitative results of our method for grasp accuracy and segmentation. The grasp accuracy is calculated using the Jaccard index, the instance seg. accuracy is reported using the mean IOU, both is done for all detected graspable objects in the scene. Foreground / background segmentation is evaluated by calculating the mean IOU of predicted and ground truth segmentation.
To evaluate the effect of depth-aware CoordConv, we trained our method with and without using these additional feature maps. By comparing the results, one can see that using depth-aware CoordConv leads to an improved overall performance, especially for the task of instance segmentation. Using depth-aware CoordConv also has a positive effect on grasp accuracy (as the grasp accuracy is indirectly influenced by depth-aware CoordConv through the shared backbone). 

Although not directly comparable, we report the results of our quantitative evaluation together with the results of \cite{ainetter2021}, to emphasize the state-of-the-art accuracy of our method. The main difference is that \cite{ainetter2021} uses class information for their prediction, whereas our method has the advantage that it is class-agnostic and does not need additional information. Furthermore, as \cite{ainetter2021} only predicts sem. segmentation, they are not able to distinguish between instances of the same object class. Additional visualizations to highlight this difference are shown in the supplementary material. Figure \ref{fig:OCID_full_results} b) - d) show grasp candidates for multiple graspable objects in the scene, and the corresponding instance segmentation and fore-/background segmentation. Figure \ref{fig:OCID_full_results} e) - f) show grasp candidate and corresponding segmentation for one graspable object. This leads to a pixel-wise classification into background, obstacle, and graspable object providing all necessary information for robotic picking. 

\begin{table}[htb!]
\centering
\begin{tabular}{|c|c|c|c|c|c|}
\hline
\multirow{2}{*}{Method}  & \begin{tabular}[c]{@{}c@{}}Grasp\\ Accuracy\end{tabular}  & \begin{tabular}[c]{@{}c@{}}Sem.\\ Seg. IOU\end{tabular} & \begin{tabular}[c]{@{}c@{}}Grasp\\ Accuracy\end{tabular} & \begin{tabular}[c]{@{}c@{}}Fore-/Background\\ Seg. IOU\end{tabular} & \begin{tabular}[c]{@{}c@{}}Inst.\\ Seg. IOU\end{tabular} \\ \hline
\begin{tabular}[c]{@{}c@{}}Grasp Detection \\ \& Sem. Seg. \cite{ainetter2021}\end{tabular}  & 89.02  & 94.05   & -  & -   & -   \\ \hline
\begin{tabular}[c]{@{}c@{}}(ours) without \\ depth-aware CC \end{tabular} & -  & -   & 95.95  & 96.65  & 92.37  \\ \hline
\begin{tabular}[c]{@{}c@{}}(ours) with \\depth-aware CC \end{tabular}   & -  & -   & \textbf{97.09}  & \textbf{97.16}  & \textbf{94.55}  \\ \hline
\end{tabular}
\vspace{5pt}
\caption{Results for grasp accuracy, semantic segmentation and instance segmentation for OCID\_grasp in (\%). Note that the method from \cite{ainetter2021} and (ours) are not comparable, as \cite{ainetter2021} uses class-specific grasp detection and segmentation. Experiments show that using depth-aware CoordConv (CC) improves overall performance of our network, especially for inst. segmentation.}
\label{tab:OCID_results}
\end{table}

\vspace{-6pt}
\subsection{Real-World Robotic Picking}
\begin{figure}[htb!] 
      \centering
 		\includegraphics[trim={0cm 0 0cm 0cm},width=0.9\linewidth]{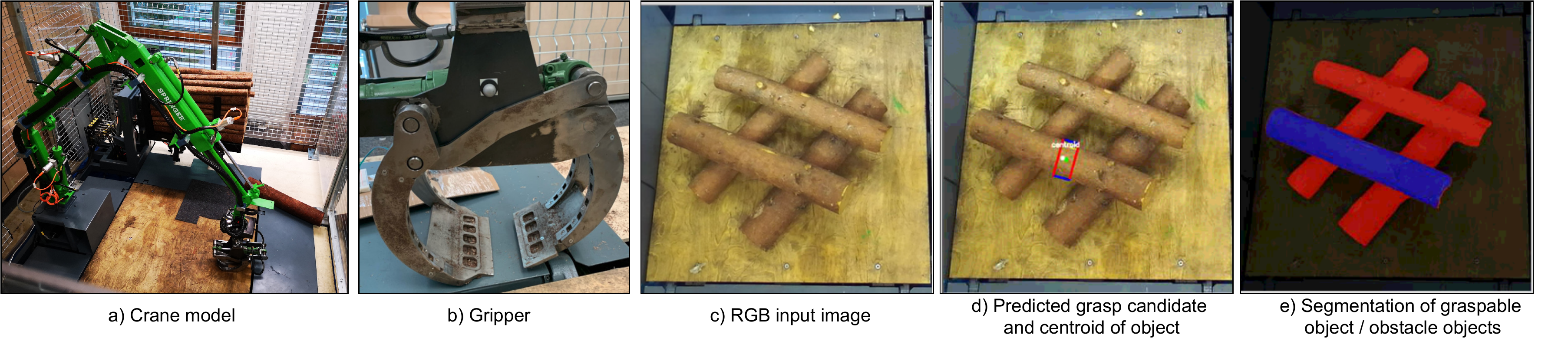}
 		\vspace{-6pt}
	\caption[]{Overview of the experimental setup and visualization of results. a) shows the 1:5 scaled version of a forestry crane, b) shows the used gripper. c) shows an example RGB input image, with the corresponding results shown in d) - e).}
	\label{fig:AL_overview}
\end{figure}

\vspace{-6pt}
\textbf{Model description and setup.} To test the accuracy of our proposed architecture for picking objects in practice, we used a hydraulic actuated crane model build for picking logs in the wood sector. This crane is a 1:5 scaled version of a real forestry crane, and the hydraulics are specifically designed to match this scale. Figure \ref{fig:AL_overview} a) and b) show the 1:5 scaled crane model and gripper, respectively. A ceiling mounted ZED camera captures RGB-D images. 
\\\textbf{Data preparation.}  We used scaled versions of real tree logs as objects in our experiments. Figure \ref{fig:AL_overview} c) shows an example of the RGB input data. Note that although the objects have no complex shape (unlike the objects in the previous experiments), this experiment serves as demonstration of how to use our method to achieve high grasp success rate.
\\\textbf{Training details.} We manually annotated 266 images (similar to the one in Figure \ref{fig:AL_overview} c)) with ground truth grasp candidates and segmentation masks. Extensive data augmentation ensured that no overfitting during training occurred. The network was trained end-to-end using a learning rate of 0.005 with weight decay of 0.0001, a momentum factor of 0.9 with enabled nesterov momentum and SGD as optimizer. We used a batch size of 2 and weighting factors of $\lambda_{grasp}= 1.0$, $\lambda_{sem}=0.8$ and $\lambda_{inst}=0.8$  for loss balancing. During training, we randomly sampled point proposals for object instance segmentation, during test runs, we used the center of the grasp candidates $\mathbf{g}$ as point proposals $p(x,y)$.
\\\textbf{Using segmentation results to increase grasp success rate for picking and placing logs.}
This robotic picking experiment is used to demonstrate simple ways to utilize the predicted segmentation to achieve a high grasp success rate. This is done by implementing three segmentation-based sanity checks: 1) checking if the inst. seg. mask is continuous (if not, it indicates that the object may be occluded, and we skip the corresponding grasp candidate), 2) optimizing the opening width of the gripper by expanding the height of the predicted grasp candidate towards the edges of the inst. segmentation mask, and 3) calculating the centroid of the graspable object using the inst. segmentation mask, as the distance from the centroid to the center of the grasp candidate is important for correctly placing the object in the storage container. Although very simple, these checks help to achieve a high success rate for picking and placing objects. Another way to leverage segmentation results is to focus on obstacle awareness and collision avoidance, similar as \cite{murali20206} proposed for 6-DOF grasping.
\\\textbf{Grasping experiments and results.}
Multiple logs were randomly placed in the scene, with different locations and orientations, often resulting in highly cluttered scenes. Our system automatically predicts a grasp candidate for a graspable log and starts the picking process. This procedure continues until no logs are detected. The grasping experiments collected data from 12 test runs with up to 7 logs each. Figure \ref{fig:AL_overview} d) - e) show an example for grasp detection and object/obstacle segmentation, respectively. Each grasp attempt is evaluated for its success and the overall result is presented in Table \ref{tab:AutoLog_grasp_eval} for completeness.  
Additional information about our robotic system can be found in the supplementary material.
\vspace{-6pt}
\begin{table}[htp!]
\centering
\vspace{-12pt}
\begin{tabular}{|c|c|c|c|}
\hline
Total attempts & Inaccurate grasp detection & Physical failure & Grasp success rate (\%) \\ \hline 
58&      2    &    1    &  94.82                                                        \\ \hline
\end{tabular}
\vspace{6pt}
\caption{Results for picking real objects in highly cluttered scenes. High grasp success rate was achieved by leveraging segmentation results to optimize picking and placing of objects.}
\vspace{-6pt}
\label{tab:AutoLog_grasp_eval}
\end{table}
\vspace{-18pt}
\section{Conclusion}
In this work, we introduced a novel CNN architecture for joint grasp detection and highly accurate class-agnostic instance segmentation for robotic picking tasks. We showed that our proposed depth-aware CoordConv module is a simple yet effective way to enhance point proposal based object instance segmentation, especially in scenes where objects highly overlap. Using our proposed network architecture, we showed highly accurate results for the challenging Siléane and OCID\_grasp datasets, and implemented a practical experiment to highlight the benefits of joint grasp detection and segmentation for picking objects. In the future, we plan to use the segmentation results for collision avoidance during path planning and picking of objects.
\section*{Acknowledgments}
\vspace{-6pt}
The research leading to these results has received funding from the Austrian Ministry for Transport, Innovation and Technology (BMVIT) within the ICT of the Future Programme (4th call) of the Austrian Research Promotion Agency (FFG) under grant no. 864807.
\vspace{-12pt}
\section{Supplementary Material}
\subsection*{Additional Analysis of Depth-aware CoordConv}
In addition to the results in Section 5.1, we report the performance using additional methods to calculate feature maps for depth-aware CoordConv.  Hereafter, we describe other configurations for depth-aware CoordConv, resulting in different feature maps.
\\\textbf{Depth similarity map $\mathbf{D}_{sim}$.} It is based on the idea of \cite{wang2018depth} to calculate the similarity of the depth image related to a specific point (in our case the point proposal $p(x,y)$). The depth similarity $\mathbf{D}_{sim}$ is defined as
\begin{equation}
\begin{split}
\mathbf{D}_{sim} =  \exp( \beta \cdot | \mathbf{D} - \mathbf{D}(p)|) - 1, \\
\mathbf{D}_{sim} \in [0,\exp(\beta) - 1],
\label{eq:depth_sim_supp}
\end{split}
\end{equation}
with the normalized depth image $\mathbf{D} \in [0,1]$, the point proposal $p$ and a scaling factor $\beta$.
\\\textbf{HHA map $\mathbf{H}_{dist}$.} HHA encoding \cite{gupta2014learning} represents the depth image by three channels (horizontal disparity, height above ground, and angle with gravity). 
Fore each channel $c \in \{1,2,3\}$ of the HHA encoding $\mathbf{H}$ we calculate a distance map as
\begin{equation}
\begin{split}
\mathbf{H}_{dist}^{c} = \alpha \cdot (\mathbf{H}^{c} - \mathbf{H}^{c}(p)), \\
 \mathbf{H}_{dist}^{c} \in [-\alpha,\alpha],
\label{eq:HHA_dist}
\end{split}
\end{equation}
whereas $\mathbf{H}^{c} \in [0,1]$ is the c-th normalized channel of the HHA encoding. \\\\
Table \ref{tab:sileane_eval_supp} shows the results for the additional experiments, whereas the setup of the experiments is identical to the one in Section 5.1. As reported in the main paper, using the depth-distance map $\mathbf{D}_{dist}$ together with the 2.5D distance map $\mathbf{F}_{2.5D}$ achieves the highest accuracy. 
Using $\mathbf{D}_{sim}$ instead of $\mathbf{D}_{dist}$ performs reasonable well, whereas the disadvantage of $\mathbf{D}_{sim}$ is that this similarity measure is not able to differentiate between positive and negative distances in z-direction, which results in an information loss. The HHA distance maps $\mathbf{H}_{dist}$ has the limitation that the HHA encoding already encode properties of geocentric pose, which leads to ambiguities when relating it to the point proposal $p$.
\begin{table}[]
\centering
\begin{tabular}{|c|c|c|c|c|c|c|}
\hline
Method                                                       & \begin{tabular}[c]{@{}c@{}}Rel. Coord-\\ Conv \cite{sofiiuk2019adaptis}\end{tabular} & \begin{tabular}[c]{@{}c@{}}Depth-dist.\\ Map\end{tabular} & \begin{tabular}[c]{@{}c@{}}2.5D Dist.\\ Map\end{tabular} & \begin{tabular}[c]{@{}c@{}}Depth-sim.\\ Map\end{tabular}& \begin{tabular}[c]{@{}c@{}}HHA \\ Map\end{tabular} & \begin{tabular}[c]{@{}c@{}}Inst.\\ Seg. IOU\end{tabular} \\ \hline
\begin{tabular}[c]{@{}c@{}}Backbone + \\ Seg. Branch \end{tabular} & \xmark   & \xmark  & \xmark  & \xmark & \xmark & 83.01 \\ \hline
\begin{tabular}[c]{@{}c@{}}Backbone + \\ Seg. Branch \end{tabular} & \cmark & \xmark & \xmark  & \xmark  & \xmark & 85.63 \\ \hline
\begin{tabular}[c]{@{}c@{}}Backbone + \\ Seg. Branch \end{tabular} & \cmark & \cmark & \cmark  & \xmark  & \xmark & \textbf{91.27}\\ \hline
\begin{tabular}[c]{@{}c@{}}Backbone + \\ Seg. Branch \end{tabular} & \cmark & \xmark & \begin{tabular}[c]{@{}c@{}}\cmark (only \\ $\mathbf{X}_{rel}$,$\mathbf{Y}_{rel}$)\end{tabular}  & \cmark  & \xmark & 90.91\\ \hline
\begin{tabular}[c]{@{}c@{}}Backbone + \\ Seg. Branch \end{tabular} & \cmark & \xmark &   \begin{tabular}[c]{@{}c@{}}\cmark (only \\ $\mathbf{X}_{rel}$,$\mathbf{Y}_{rel}$)\end{tabular}  &  \xmark  & \cmark & 89.68   \\ \hline
\end{tabular}
\caption{Comparison of different configurations of depth-aware CoordConv for instance segmentation on Siléane dataset. Setup and execution of experiments are identical to the ones in Section 5.1 in the main paper.}
\label{tab:sileane_eval_supp}
\end{table}
\subsection*{Detailed Information about Loss Function}
For simultaneously learning the tasks of grasp detection, semantic segmentation and instance selection, we defined the composite loss function as
\begin{equation}
\mathcal{L}= \lambda_{grasp}\mathcal{L}_{grasp} + \lambda_{sem}\mathcal{L}_{sem} + \lambda_{inst}\mathcal{L}_{inst},
\label{eq:LOSS_full}
\end{equation}
with the grasp detection loss $\mathcal{L}_{grasp}$, the semantic segmentation loss $\mathcal{L}_{sem}$, and the instance selection loss $\mathcal{L}_{inst}$. All parts are weighted with a specific hyperparameter $\lambda$.
\\\textbf{Grasp detection loss.} The grasp detection loss $\mathcal{L}_{grasp}$ is defined as
\begin{equation}
\mathcal{L}_{grasp}= \mathcal{L}_{RPN} + \mathcal{L}_{box} + \mathcal{L}_{rot},
\label{eq:LOSS_grasp}
\end{equation}
where $\mathcal{L}_{RPN}$ defines the loss for training a Region Proposal Network (RPN), which is part of the grasp detection branch. $\mathcal{L}_{box}$ defines the regression loss for the box coordinates $(x,y,w,h)$ and $\mathcal{L}_{rot}$ defines the classification loss for the grasp orientation $\theta$.
We refer to \cite{he2017mask} for additional information about the RPN and corresponding loss $\mathcal{L}_{RPN}$. 
The grasp orientation loss $\mathcal{L}_{rot}$ is defined as
\begin{equation}
\mathcal{L}_{rot} = - \frac{1}{\mathcal{|R|}} \sum_{r \in \mathcal{R_{+}}} \log s_{r}^{c_{\theta}}   - \frac{1}{\mathcal{|R|}} \sum_{r \in \mathcal{R_{-}}} \log s_{r}^{c_{\emptyset}}.
\label{eq:rot_loss}
\end{equation}
Note that $\mathcal{R} = \mathcal{R_{+}} \cup \mathcal{R_{-}}$ is the set of valid and invalid region proposals, which are the output of the RPN. Each region proposal $r$ consists of the parameters $(x_{r},y_{r},w_{r},h_{r})$, which represent an initial axis-aligned bounding box. The score function $s_{r}^{c_{\theta}}$ defines the probability that the region proposal belongs to the ground truth orientation class $c_{\theta}$, and $s_{r}^{\emptyset}$ defines the probability that the region proposal is invalid. Note that we discretized the grasp orientation $\theta$ into $18$ intervals with equal length, where each interval is represented by its mean value, resulting in $c_{\theta} \in \{1,...,18\}$ for the orientation classes. The additional class $c_{\emptyset}$ is used to describe the region proposals which may be invalid.  \\
For bounding box regression we use the loss $\mathcal{L}_{box}$ defined as
\begin{equation}
\mathcal{L}_{box} = \sum_{i \in \{x,y,w,h\}} smooth_{L_{1}} (t_{i} - t_{i}^{*}), 
\label{eq:box_loss}
\end{equation}
with the $smooth_{L_{1}}$ norm defined in \cite{ren2015faster}. The correction factors $t_{i}$ are calculated by the grasp detection head, and $t_{i}^{*}$ represent the offset between ground truth grasp candidates and a region proposal $r$. The correction factors $t_{i}$ and the corresponding region proposal parameters $(x_{r},y_{r},w_{r},h_{r})$, are then used to calculate the final box parameters $(x,y,w,h)$.
\\\textbf{Semantic segmentation loss.} We denote $l = \{1,...,N\}$ as the set of semantic segmentation classes, with $N=2$ for foreground/background segmentation. The semantic segmentation loss is a weighted per-pixel loss \cite{porzi2019seamless} defined as
\begin{equation}
\mathcal{L}_{sem} = - \sum_{j,k} w_{j,k} \log P_{j,k}(Y_{j,k}),
\label{eq:sem_loss}
\end{equation}
where $(j,k)$ correspond to the pixel position in the image. Let $Y_{j,k} \in l$ be the semantic segmentation ground truth and $P_{j,k}$, the predicted probability of the semantic segmentation head for the same pixel, to be assigned to one of the semantic classes, respectively. The weights $w_{j,k}$ select the $25\%$ of the lowest predicted probabilities $P_{j,k}$ for all $(j,k)$ using $w_{j,k} = \frac{4}{WH}$, and $w_{j,k} = 0$ otherwise, whereas $(W \times H)$ is the spatial image resolution.
\\\textbf{Instance segmentation loss.} The instance segmentation is defined as binary segmentation problem, given that we want to calculate a specific object instance mask related to the point proposal $p(x,y)$. We use the Normalized Focal Loss as proposed in \cite{sofiiuk2019adaptis}, which is defined as
\begin{equation}
    \mathcal{L}_{inst} = -\frac{1}{Q(M)} \sum_{j,k} (1 - Q_{j,k})^{\gamma} \log Q_{j,k},
\end{equation}
whereas $Q(M) = \sum_{j,k} (1 - Q_{j,k})^{\gamma}$ defines the total weight of the values for all pixels in the image, with $Q_{j,k}$ the predicted probability of the instance selection network for the correct segmentation output at position $(j,k)$, and the focusing parameter $\gamma \geq 0$. The Normalized Focal Loss concentrates on pixels that are misclassified by the network, in contrast to the binary cross entropy loss, which pays more attention to pixels that are correctly classified (as shown in \cite{lin2017focal}).
\subsection*{Additional Visualizations for OCID\_grasp Experiment}
\begin{figure}[] 
      \centering
 		\includegraphics[trim={0cm 0cm 0cm 0},width=\linewidth]{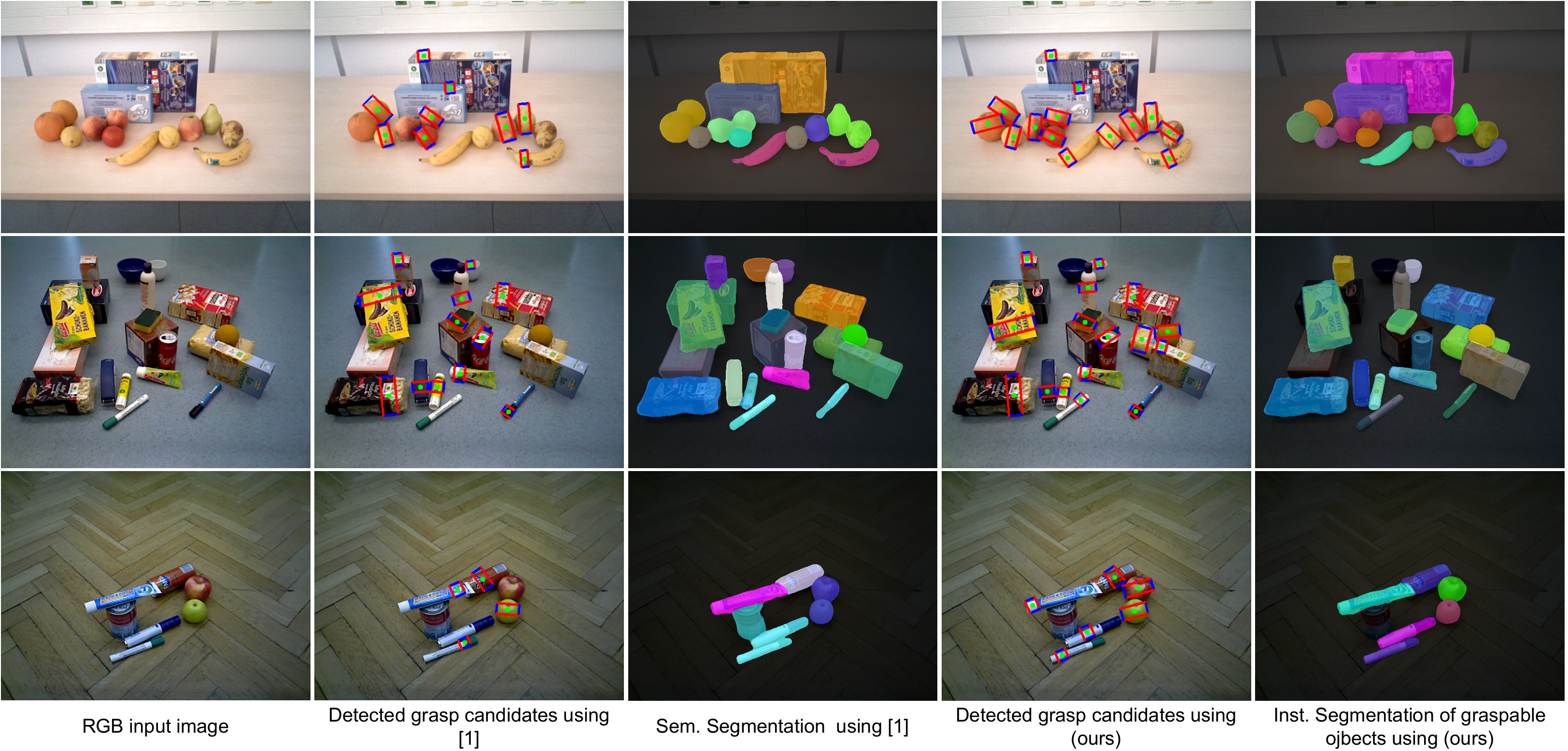}
 		\caption{Comparison with previous state-of-the-art method for joint grasp detection and segmentation. As the method of \cite{ainetter2021} selects grasp candidates related to the semantic segmentation, they are missing out on grasp candidates if multiple instances of the same object class are present in the scene (e.g. for classes banana and orange in row one, column two). Furthermore, the semantic segmentation in \cite{ainetter2021} makes it impossible to distinguish between multiple object instances of the same class, whereas (ours) performs class-agnostic instance segmentation. Failure cases for (ours) are if no grasp candidate is detected for a graspable object, see dark bowl in second row, fourth column. Note that the colors for the seg. results in \cite{ainetter2021} and (ours) are independent and have no correlation.}
	\label{fig:ICRA_BMVC_compare2}
\end{figure}

Figure \ref{fig:ICRA_BMVC_compare2} shows comparison of qualitative results for \cite{ainetter2021} and (ours), corresponding to Section 5.2 in the paper. Please note the differences: 1) \cite{ainetter2021} performs semantic segmentation, where each color in the visualization corresponds to a certain object class, whereas multiple object instances of the same class share the same color. This makes it impossible to distinguish between multiple instances of the same class. (Ours) performs class-agnostic instance segmentation for multiple graspable objects, where each instance has a separate color, not related to classes. 2) Because \cite{ainetter2021}
filters grasp candidates using the semantic segmentation, it is only possible to select one grasp candidate per class. Again, if multiple instance of one object class are in the scene, this algorithm fails to detect grasp candidates for more than one instance. (Ours) on the other hand, is free of restrictions regarding object classes, and is therefore able to predict grasp candidates for multiple instances of the same class.
Failure cases for our method are if no grasp candidate is predicted for a graspable object (as one can see in Figure \ref{fig:ICRA_BMVC_compare2} second row, fourth column, no grasp candidate for bowl detected), which then results in no instance segmentation mask, as the center of the grasp candidate is used as point proposal for instance segmentation.
\begin{figure}[]
    \centering
    \includegraphics[trim=85 0 100 0, clip, width=0.4\linewidth]{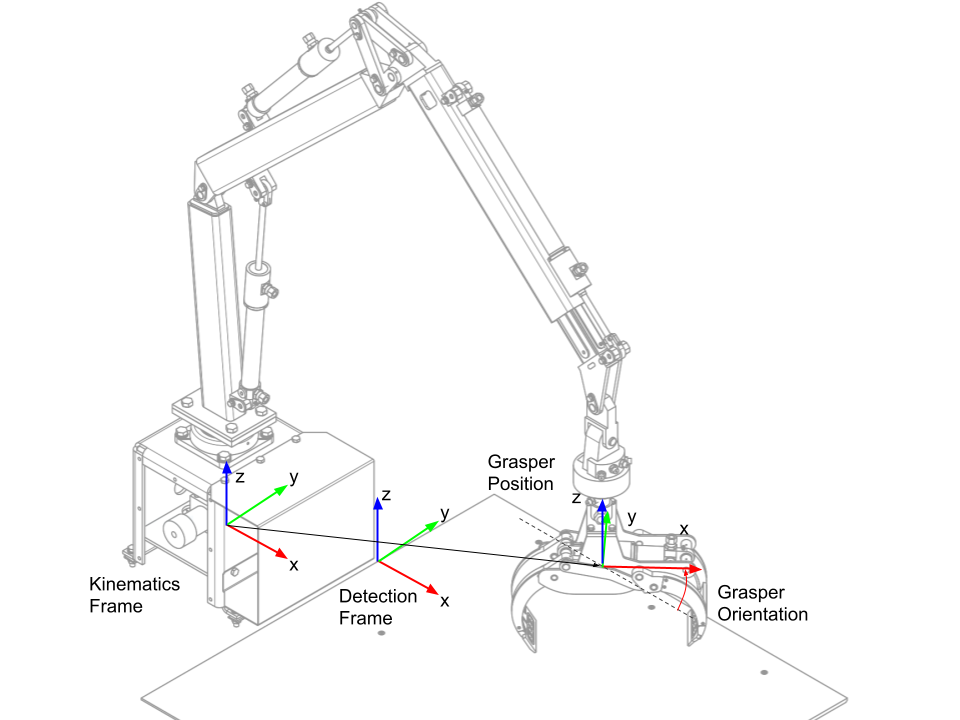}
    \quad
    \includegraphics[width=0.48\linewidth, trim=0 380 1200 0, clip]{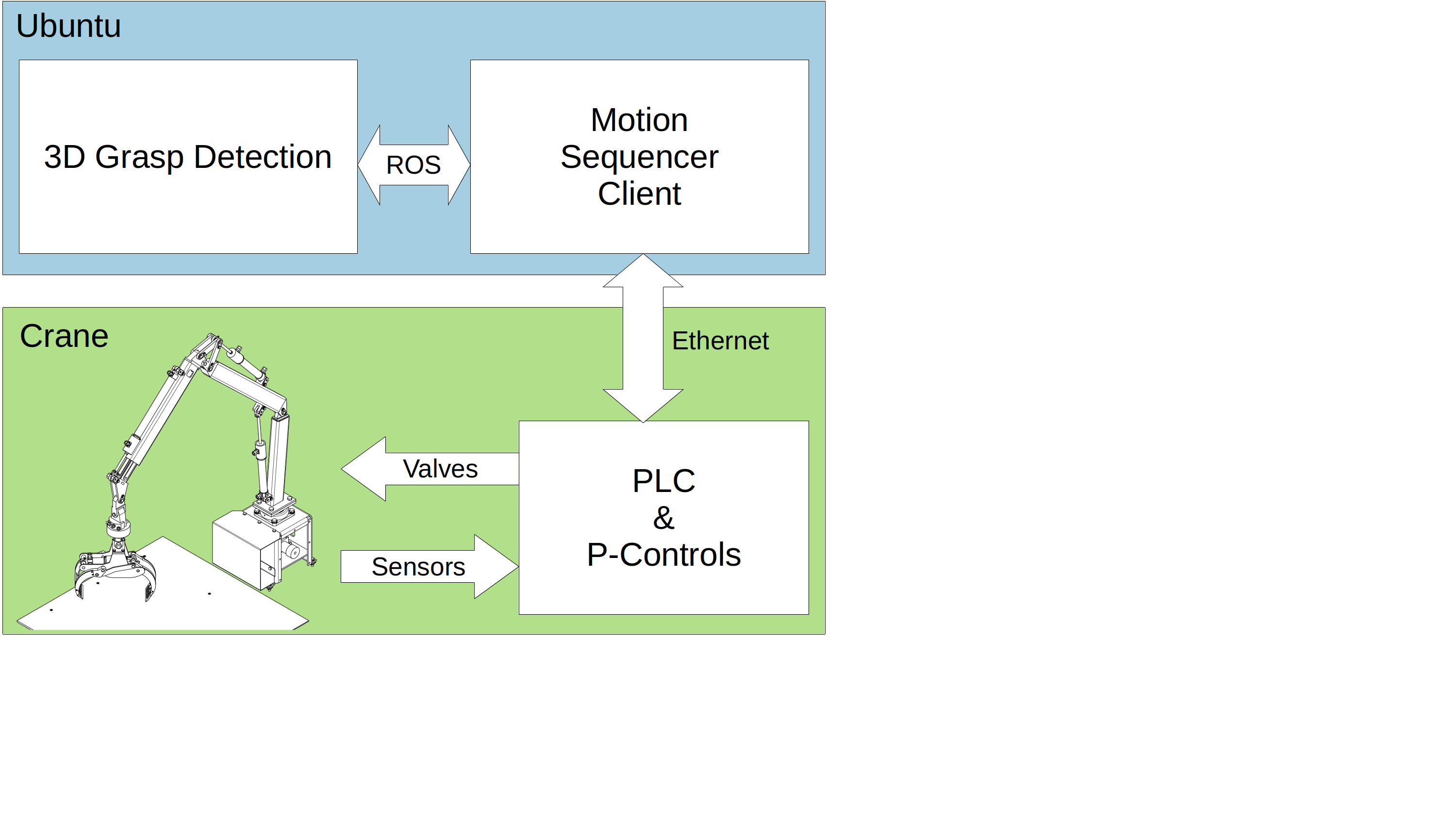}
 	\caption{The drawing, left image, shows the crane used in real-world experiments and important reference frames; An overview of the system structure can be seen in the right image.}
 	\label{fig:system_overview}
\end{figure}
\subsection*{Technical Details for Real-World Robotic Picking}
The system used for the real-world robotic picking experiments is a scaled hydraulic forest crane (see left-hand drawing in Figure \ref{fig:system_overview}). These cranes usually are human-operated to move tree logs from one place to another. To allow for automatic pick-and-place operation, we added electrically operated hydraulic valves to each joint of the crane in combination with a programmable logic controller (PLC). Linear or angular displacement sensors with electrically operated hydraulic valves make the automatic control of the crane possible.

The right-hand diagram of Figure \ref{fig:system_overview} shows the overall structure of the system. The top left block includes the detection of feasible grasp candidates (see method in Section 4 in main paper). The best grasp candidate is projected into 3D and sent to the so-called motion sequencer client ( Figure \ref{fig:system_overview} top right) through a ROS message. This message contains the grasping 3D position, its orientation, the log's diameter and centroid. With this data (transformed from detection to the kinematics frame) and the periodically arriving sensor measurements from the PLC, the motion sequencer client can calculate the configuration of the joints needed to reach the requested point. A minimization method uses forward kinematics of the crane and the desired position and orientation to calculate the required joint configuration.

The PLC receives the calculated joint configuration via Ethernet. On the PLC, each joint has a proportional control loop with the received configuration as desired position/orientation, the measurements as current position/orientation, and the valve opening as output. The control loop interpolates linearly between the old and new desired setpoint, to avoid sudden motions and unwanted pressure spikes during setpoint changes. The motion sequencer client checks through the reported configuration if it matches the new setpoint. One needs to be aware that the hydraulics combined with a proportional control loop limits the precision in positioning, compared to just electrical joint-based systems.

Always heading directly for the detected grasping pose might result in erroneous behavior or worst-case crashes of the crane with its surroundings. Therefore, the motion sequencer client generates a sequence of so-called waypoints, based on the received grasp candidate, to ensure repeatability and safety. This sequence guarantees that side-ways motions are only done while high above the ground and only going down to pick up a log or put it into the container. The grasper itself limits grasping complexity as it is bulky because of its application, compared to other pick-and-place robotic manipulators. Thus, one needs to take care of the distance between logs. The grasper only opens as wide as the log's diameter plus safety margin to improve grasping performance in cluttered scenarios. A ROS service on the grasping detection, which triggers a prediction, allows the automation of the experiment. The motion sequencer client calls this service to start a new sequence. This procedure continues until no grasp candidates are detected.

\bibliography{egbib}
\end{document}